\documentclass[conference]{IEEEtran}
\IEEEoverridecommandlockouts
\usepackage{cite}
\usepackage{amsmath,amssymb,amsfonts}
\usepackage{subcaption}
\usepackage{algorithmic}
\usepackage{graphicx}
\usepackage{textcomp}
\usepackage{xcolor}
\usepackage{booktabs}
\usepackage[bookmarks=false, hidelinks, pdfborder={0 0 0}]{hyperref}
\usepackage[T1]{fontenc}

\def\BibTeX{{\rm B\kern-.05em{\sc i\kern-.025em b}\kern-.08em
    T\kern-.1667em\lower.7ex\hbox{E}\kern-.125emX}}
\begin{document}

\title{RSEND: Retinex-based Squeeze and Excitation Network with Dark Region Detection for Efficient Low Light Image Enhancement\\
}

\author{\IEEEauthorblockN{1\textsuperscript{st} Jingcheng Li}
\IEEEauthorblockA{\textit{Department of Computer Science and Engineering} \\
\textit{University of California, San Diego}\\
La Jolla, United States \\
jil458@ucsd.edu}
\and
\IEEEauthorblockN{2\textsuperscript{nd} Ye Qiao}
\IEEEauthorblockA{\textit{Department of Electrical Engineering and Computer Science} \\
\textit{University of California, Irvine}\\
Irvine, United States \\
yeq6@uci.edu}
\and
\IEEEauthorblockN{3\textsuperscript{rd} Haocheng Xu}
\IEEEauthorblockA{\textit{Department of Electrical Engineering and Computer Science} \\
\textit{University of California, Irvine}\\
Irvine, United States \\
haochx5@uci.edu}
\and
\IEEEauthorblockN{4\textsuperscript{th} Sitao Huang}
\IEEEauthorblockA{\textit{Department of Electrical Engineering and Computer Science} \\
\textit{University of California, Irvine}\\
Irvine, United States \\
sitaoh@uci.edu}
}

\maketitle

\begin{abstract}
 Images captured under low-light scenarios often suffer from low quality. Efficient low-light image enhancement with mobile computing has become an urgent need. Previous CNN-based low-light image enhancement methods often involve using Retinex theory. Nevertheless, most of them do not perform well in complicated datasets like LOL-v2 while using too much computational resources. Besides, some of these methods require sophisticated training at different stages, making the procedure even more time-consuming and tedious. In this paper, we propose an accurate, concise, and one-stage Retinex theory-based framework with a novel dark region detection module and Squeeze and Excitation blocks for enhanced detail retention, RSEND, for efficient low-light image enhancement. RSEND first divides the low-light image into the illumination map and reflectance map, then detects the different dark regions in the illumination map and performs light enhancement. After this step, it refines the enhanced gray-scale image and does element-wise matrix multiplication with the reflectance map. By denoising the output it has from the previous step, it obtains the final result. In all the steps, RSEND utilizes Squeeze and Excitation network to better capture the details. Comprehensive quantitative and qualitative experiments show that our efficient Retinex model significantly outperforms other CNN-based state-of-the-art models, achieving a PSNR improvement ranging from 1.69 dB to 3.63 dB in different datasets. Compared to Transformer-based models, RSEND achieves higher PSNR values ranging from 1.22 dB to 2.44 dB in the LOL-v2-real dataset. Importantly, RSEND achieves these performance improvements with remarkable efficiency, utilizing only 0.41 million parameters, which represents a substantial reduction (3.93--9.78$\times$) in computational resources compared to existing state-of-the-art methods. The code can be found at \href{https://github.com/jeffconqueror/RSEND/tree/main} {https://github.com/jeffconqueror/RSEND/tree/main}.
\end{abstract}

\section{Introduction}
\noindent Low-light image enhancement aims to improve the visibility and perceptual quality of underexposed images captured in underlit environments. This enhancement problem is challenging as it requires careful handling of noise, color distortion, loss of details, etc. while keeping the model training and inference computing footprint reasonably low. 

\begin{figure*}[t]
    \centering
    \includegraphics[width=1\textwidth]{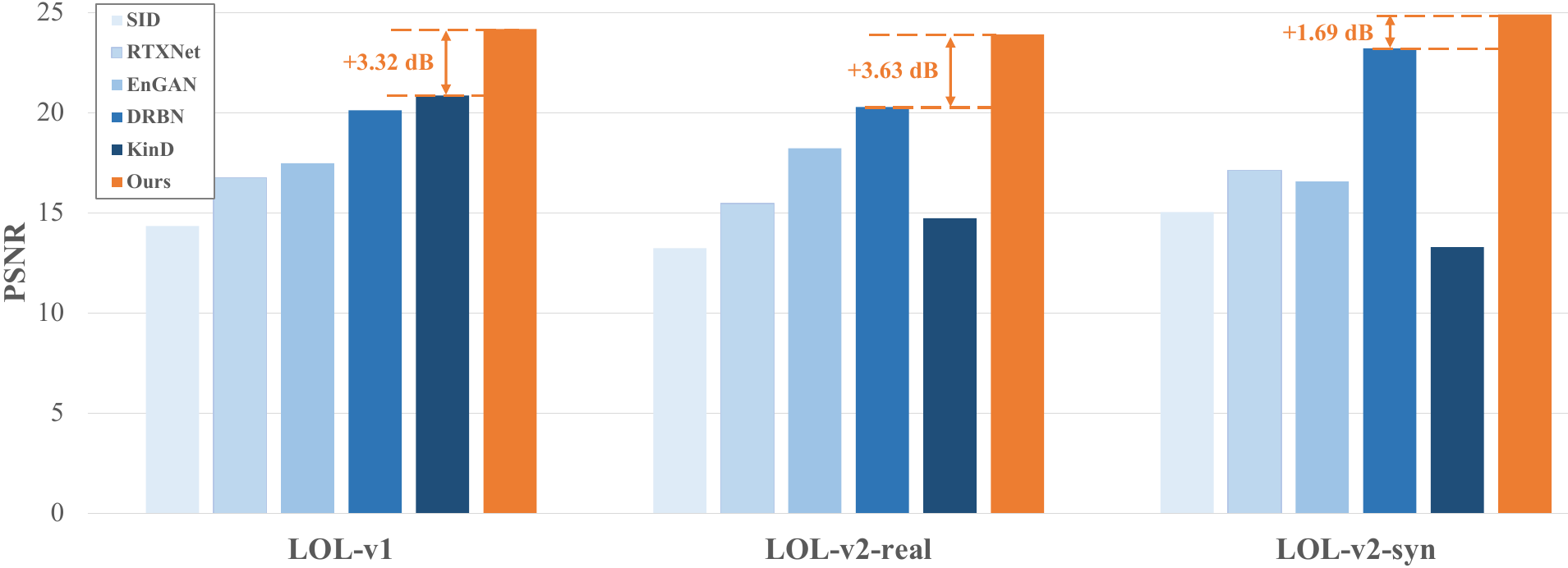} 
    \caption{Comparison of our \textbf{RSEND} against previous CNN based state-of-the-art methods including \textbf{SID} \cite{chen2019seeing}, \textbf{RetinexNet (RTXNet)} \cite{wei2018deep}, \textbf{EnGAN} \cite{jiang2021enlightengan}, \textbf{DRBN} \cite{yang2021band}, \textbf{KinD} \cite{zhang2019kindling} on three datasets: {LOL-v1}, {LOL-v2-real}, and {LOL-v2-syn}. Our RSEND flow achieves 1.69 dB to 3.63 dB improvements over the best previous works in terms of PSNR, as indicated by the  {\color{orange} orange +dB} annotations.}
    \label{fig:my_label} 
\end{figure*}

\noindent Many prior approaches have been proposed for low-light image enhancement in literature. Previous image enhancement methods mainly include Retinex theory-based \cite{land1971lightness} \cite{wei2018deep} algorithm and histogram equalization \cite{ibrahim2007brightness}. However, both techniques have their own drawbacks. Histogram equalization sometimes leads to over-amplification of noise in relatively dark areas of an image as well as a loss of detail in brighter sections \cite{cai2023retinexformer}.  This technique applies a global adjustment to the image's contrast, which may not be suitable for images where local contrast variations are important for detail visibility. As for Retinex theory-based methods, they assume an image could be decomposed into illumination and reflectance, while preserving the reflectance, by only enhancing the illumination can they get the final enhanced image. However, such methods sometimes produce results that appear unnatural due to over-enhancement or incorrect color restoration. Additionally, these models might struggle with very dark regions where information is minimal, potentially leading to artifacts or noise amplification.

\noindent With the advancement of deep learning, convolutional neural networks (CNNs) have become a pivotal technology for enhancing images captured in low-light conditions. Various types of CNN-based architectures have been explored. They are mainly divided into two categories, generative adversarial networks (GANs) \cite{jiang2021enlightengan} and models inspired by Retinex theory \cite{wei2018deep} \cite{liu2021retinex} \cite{cai2023retinexformer}. For GANs, the adversarial process helps improve the quality of the enhancement, making images look more natural. However, GAN-based methods suffer from training instability, leading to artifacts or unrealistic results, especially under complex lighting conditions \cite{tian2019deep}. As for Retinex theory-based deep learning models, they typically decompose an image into reflectance and illumination components. The models estimate these components separately, enhancing the illumination part to improve image visibility while preserving the reflectance to maintain color fidelity and details. One drawback of such models is their reliance on accurate decomposition, which can be challenging in complex lighting conditions, and they always suffer from the multi-stage training pipeline. 

\noindent Another problem is that previous methods typically require building models with a large number of parameters, which leads to heavy computational complexity that is unaffordable in certain situations, e.g., image enhancement on mobile devices. Besides computational considerations, privacy concerns are paramount when processing sensitive images on local devices. If we have a compact and efficient network, we can ensure that image processing occurs locally on the device itself, especially for mobile applications where memory and processing power are limited. Not only for mobile and edge devices, reducing computing cost is also a critical need for deploying models in the cloud. Smaller models require less computational power, thereby saving more energy and reducing the financial burden associated with cloud resources. 

\noindent To fix the aforementioned problems, we propose a novel method, RSEND, for efficient low-light image enhancement with high quality and low computing cost. First, RSEND adopts the Retinex theory methodology and decomposes the low-light image into the illumination map and the reflectance map. Prior works didn't pay attention to locating the areas that need the most enhancement. We solve this by adding a dark region detection module so that the illumination map can further go through a multi-scale, separate pathway to locate the features that require enhancing, before the actual enhancement. Then, RSEND goes through our custom U-shape \cite{ronneberger2015u} enhancer, and refine the input image for better details. After the enhanced grayscale image does element-wise multiplication with the reflectance map, we add the original image back to maintain similarity. And finally, RSEND denoises the output for a more visually pleasing result. By utilizing Squeeze-and-Excitation Blocks \cite{hu2018squeeze} in all the steps mentioned, the network can recalibrate channel-wise feature responses by explicitly modeling inter-dependencies between channels, capturing more image details while not increasing substantial computational cost. 
Fig. \ref{fig:my_label} shows the peak signal-to-noise ratio (PSNR) result of our RSEND flow compared against state-of-the-art works on three representative datasets.  Our RSEND flow achieves 1.69 dB to 3.63 dB improvements over the best previous works. 
We will open source this work to facilitate future research. Link to the source code can be found in the supplementary materials. 

\noindent The major contributions of this work can be summarized as follows:
\begin{itemize}
    \item We propose RSEND, a one-stage Retinex-based network, for efficient low-light image enhancement with light computation and high accuracy, free from tedious multi-stage training, and maintains good performance. 
    \item Our method leverages squeeze and excitation network \cite{hu2018squeeze} to significantly enhance the representational power of the network, we largely make our network perform better without a substantial increase in computational complexity. 
    \item We use a residual learning way in the reconstruction step to make sure our output is similar to the original low-light image, maintaining high structural similarity index measure (SSIM).
    \item Our RSEND model outperforms all other CNN-based low-light image enhancement networks by up to 3.63 dB and even Transformer-based models by up to 2.44 dB while utilizing only 0.41 million parameters, which is 3.93 -- 9.78$\times$ reduction in model size. 
\end{itemize}

\section{Related Works}

\subsection{Traditional Methods}
\noindent Traditional methods for low-light image enhancement, such as histogram equalization \cite{ibrahim2007brightness} \cite{coltuc2006exact} and gamma correction \cite{farid2001blind}, focus on globally adjusting image contrasts or brightness. While these methods are simple and fast, they often overlook local context and can lead to unrealistic effects or artifacts, such as over-enhancement or under-enhancement in certain areas, or amplified noise. These limitations stem from their global processing nature, which does not account for local variations in light distribution within an image. As a result, while effective for moderate adjustments, they may struggle with images having complex light conditions or requiring nuanced enhancements.

\subsection{Deep Learning Methods}
\noindent With the fast development of deep learning, CNNs \cite{yang2021band, chen2019seeing, zhang2019kindling, fan2022multiscale} have been extensively used in low light image enhancement. EnlightenGAN \cite{jiang2021enlightengan} utilizes unsupervised learning for low-light image enhancement, leveraging a global-local discriminator structure to ensure detailed enhancement and incorporates attention mechanisms to refine areas needing illumination adjustment, but a potential drawback is the challenge of maintaining naturalness and avoiding over-enhancement, especially in images with highly variable light conditions. ZeroDCE \cite{guo2020zero} tackles low-light image enhancement through a novel deep curve estimation approach that dynamically adjusts the light enhancement of images without needing paired datasets, which introduces a lightweight deep network to learn enhancement curves directly from data. However, like many approaches using unpaired datasets, its performance may depend on the diversity and quality of the training data, potentially limiting its adaptability to unseen low-light conditions. Retinex-based deep learning models \cite{wei2018deep} focus on separating the illumination and reflectance components of an image, allowing for the manipulation of illumination while preserving the natural appearance of the scene. Nevertheless, these networks always suffer from muti-stage training pipelines and face challenges including maintaining color fidelity and avoiding artifacts.

\subsection{Squeeze-and-Excitation Network}
\noindent The Squeeze-and-Excitation Network (SENet) \cite{hu2018squeeze} introduces a mechanism to recalibrate channel-wise feature responses adaptively by explicitly modeling interdependencies between channels. It squeezes global spatial information into a channel descriptor using global average pooling, then captures channel-wise dependencies through a simple gating mechanism, and finally excites the original feature map by reweighting the channels. The SENet approach has been applied to a wide range of tasks beyond low-light image enhancement, such as image classification, object detection, semantic segmentation, and medical image analysis. 
The MLLEN-IC work (Multiscale Low-Light Image Enhancement Network with Illumination Constraint) \cite{fan2022multiscale} utilizes SENet for low-light image enhancement. The paper presents a comprehensive solution by combining a multiscale network architecture with an illumination constraint, and by using SENet, the model better restores the color and details of the image. However, it consumes substantial amount of computational resources for training and inference while only has an average PSNR value of 15.11 and SSIM value of 0.56.

\begin{figure*}[t]
    \centering
    \includegraphics[width=1.0\textwidth]{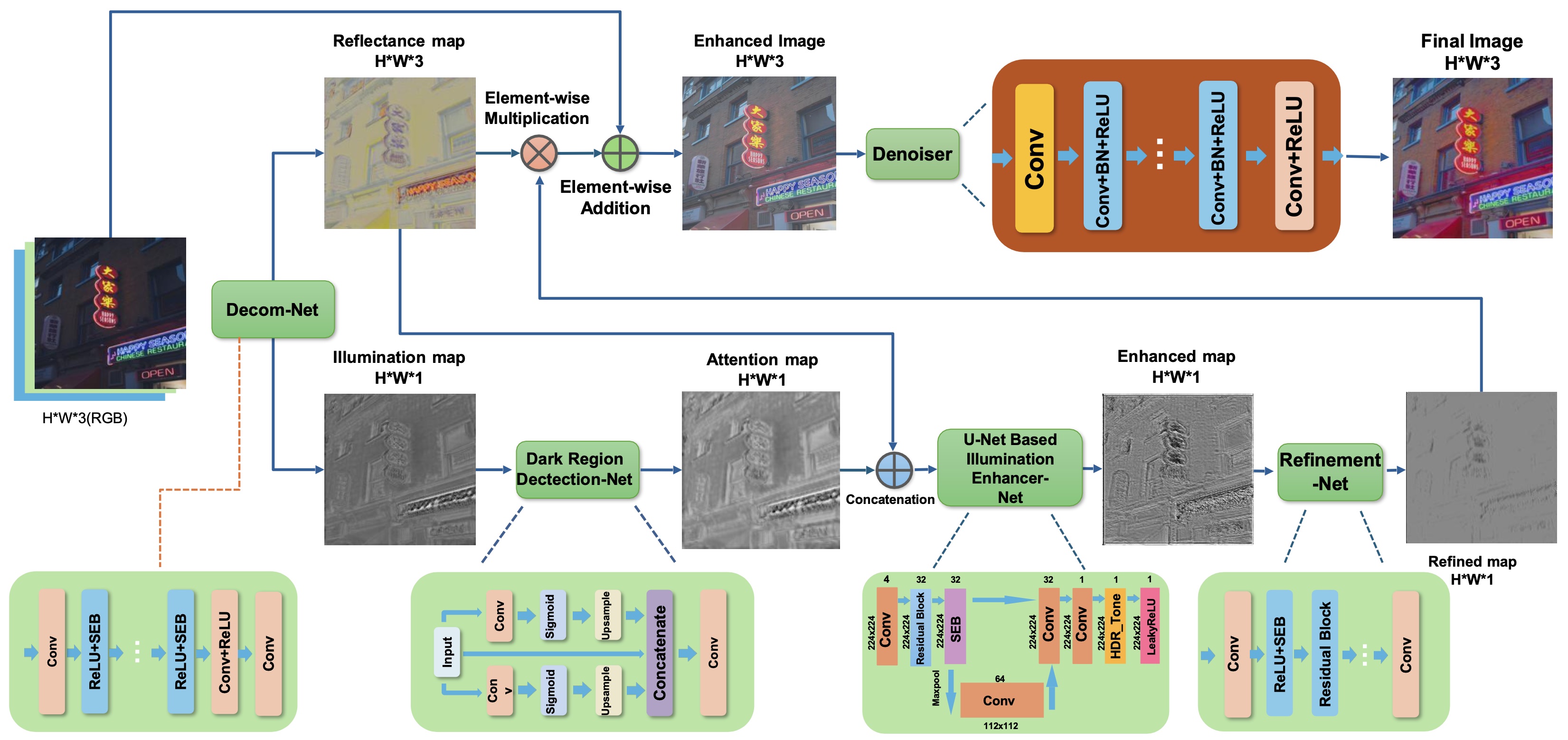} 
    \caption{\small \textbf{The proposed framework of RSEND.} Our network consists of five subnets: a Decom-Net, a Dark Region Detection-Net, an Enhancer-Net, a Refinement-Net, and a Denoiser. The Decom-Net decomposes the low-light image into a reflectance map and an illumination map based on the Retinex theory. The Dark Region Detection-Net means to find the regions that need to be enhanced more. The Enhancer-Net functions to illuminate the illumination map. The Refinement-Net aims to adjust contrasts and fine-tune the details. In the end, Denoiser performs denoising to get clean and visually pleasing output.}
    \label{fig:architecture}
\end{figure*}

\section{RSEND: Efficient Low-Light Image Enhancement}
\noindent In this section, we introduce our proposed low-light image enhancement framework, RSEND. The overall architecture of RSEND is presented in Fig. \ref{fig:architecture}. 

\subsection{End-to-end Retinex-based Model}

\noindent As we mentioned previously, RSEND first applies Retinex theory \cite{land1971lightness} to the input low-light image. A low light image $S \in \mathbb{R}^{H \times W \times 3}$ can be decomposed into reflectance $R \in \mathbb{R}^{H \times W \times 3}$ and illumination $I \in \mathbb{R}^{H \times W}$
\begin{equation}
    S = R \circ I,
\end{equation}
where the $\circ$ operator represents element-wise multiplication along the $H\times W$ dimensions and repeated across RGB channels. Similar to the previous approaches, we decompose the image into reflectance and illumination in the first step. Although many current mainstream methods have similar decomposition step, we still find there is room to improve the quality in both reflectance and illumination steps. Unlike RetinexNet \cite{wei2018deep} which manually separates the first three channels as $R$ and last channel as $I$ after a few convolutional layers, 
after feature extraction, we use two different layers in the end, one outputs three channels and the other outputs one channel, to separate $R$ and $I$ and use SEBlock \cite{hu2018squeeze} in the middle for better feature extraction. 

\subsection{Dark Region Detection Module}
\begin{figure}[t]
    \centering
    \begin{subfigure}[b]{0.45\linewidth}
        \centering
        \includegraphics[width=\linewidth]{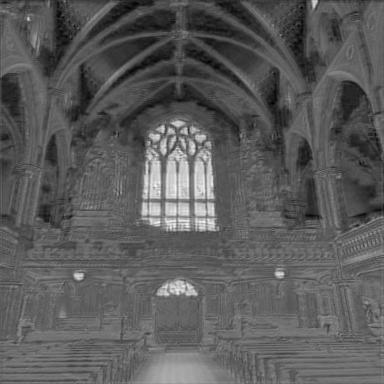}
        \caption*{Before Dark Region Detection}
    \end{subfigure}
    \hspace{0.02\linewidth}
    \begin{subfigure}[b]{0.45\linewidth}
        \centering
        \includegraphics[width=\linewidth]{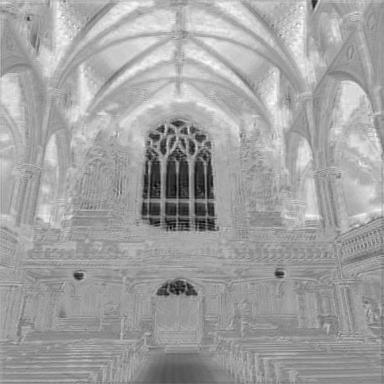}
        \caption*{After Dark Region Detection}
    \end{subfigure}
    \caption{Effect of Dark Region Detection. The left image shows the illumination map before dark region detection, while the right image demonstrates enhanced focus on darker areas after applying the module.}
    \label{fig:dark_region}
\end{figure}

\noindent Instead of directly enhancing the illumination map, we introduce a novel dark region detection module designed to emphasize areas requiring greater enhancement. As shown in Fig. \ref{fig:dark_region}, the left side presents the illumination map before dark region detection, where uniform enhancement results in limited visibility improvement. The right side illustrates the effect after applying our dark region detection module, highlighting enhanced regions more effectively. \\
\noindent We use convolutions at Illumination map with kernel 3x3 and 5x5 and stride set as 2 to capture features at different scales, then apply sigmoid activation to generate attention maps, which weights the importance of each region needs to be enhanced. By upsampling the features from different scales to the original size and doing concatenation with the original feature map, we come up with a feature map that has channels*3 depth and features from both attention-augmented pathways. In the end, we apply a 1x1 kernel size convolutional layer to the concatenated multi-scale features to reduce the channel dimensions to $I \in \mathbb{R}^{H \times W \times 1}$. By applying this module, the network is expected to pay more attention to the darker areas that need enhancement.
\begin{equation}
    \hat{I} = D(I),
\end{equation}

\subsection{Illumination Optimization}
\noindent As we illustrated in Fig. \ref{fig:architecture}, our illumination enhancer is a type of U-Net \cite{ronneberger2015u} architecture. The input of this enhancer is the pre-processed illumination map concatenated with reflectance, which gives the enhancer more details about the image's lighting and color information to consider besides the gray-scale image. For encoder, after applying convolutions to increase the channel depth to 32, we apply residual block to prevent vanishing gradient and SEBlock for modeling inter-dependencies between channels. In the bottleneck, our depth increases to 64 for capturing more complex features, and in our case, a depth of 64 is enough. As for the decoder path, the skip connections between the encoder and decoder blocks allow for the combination of high-level features.  In the final layer, we apply high dynamic range (HDR) \cite{eilertsen2017hdr} and tone mapping \cite{mantiuk2008display}, processing the feature maps from the decoder path to produce the final output image that has enhanced details in both the bright and dark areas. 
\begin{equation}
    \overline{I} = E(\hat{I}, R),
\end{equation}
After enhancement, we have a layer for refinement, which aims to fine-tune the details, adjust contrasts, and improve overall image quality. The network designed here is simpler, the core of this layer is made of convolutional layers with a kernel size of 3 and padding of 1, and like previous parts, we implement residual blocks and SEBlocks for better performance.

\subsection{Reconstruction and Denoising Phase}
\noindent As mentioned previously, after we obtain the enhanced $I$, we perform element-wise multiplication with $R$. However, unlike the Retinex theory, we add the original low-light image to the product, which serves as a form of residual learning \cite{he2016deep}, helping to retain the structure and details from the original image while adjusting the illumination. 
\begin{equation}
    \overline{S} = \overline{I} \circ R + S,
\end{equation}
Even though Retinex theory successfully enhances a low-light image, the process of enhancing can introduce or amplify noise. This is because when you increase the brightness of the dark regions, where the signal-to-noise ratio is usually lower, you also make the noise more visible. So after reconstruction, we also add a denoising phase, ensuring that the final output image is not only well-illuminated but also clean and visually pleasing. It's important to note that for real-world images, especially those taken in low-light conditions, are likely to contain noise. Thus, denoising is a crucial step in the image enhancement pipeline. Our denoising architecture is inspired by DnCNN \cite{zhang2017beyond}, it is constructed as a sequence of convolutional layers, batch normalization layers, activation functions, SEBlocks, and residual blocks. In the forward method, residual learning is applied, and the input is added back to the output of the network, ensuring that the denoised image maintains structural similarity to the original. The final formula can be formulated as
\begin{equation}
    \overline{S} = \epsilon(E(D(I), R) \circ R + S).
\end{equation}

\subsection{Compact Network}
\noindent For low-light image enhancement tasks, previous CNN-based models always add layers and increase depth for more feature representation. However, this usually does not yield promising results while largely increasing computational costs. Our RSEND framework exemplifies the principle of reducing computational costs by integrating key design choices that promote efficiency. Firstly, the use of Squeeze-and-Excitation blocks allows the model to perform dynamic channel-wise feature recalibration, which significantly boosts the representational power of the network without a proportional increase in parameters. Secondly, we carefully design the depth of the network so that each layer contributes meaningfully to the feature extraction process. Previous works always reach a depth of 512 in the bottleneck of the enhancer, while in our work, the bottleneck only has 64 channels and in other modules of the network the depth is restrained to 32. The result in Table \ref{tab:my_label} shows that it is possible to build powerful yet compact models with a fraction of the parameters in the realm of low-light image enhancement.

\section{Experiment}
\subsection{Datasets and Implementation details}
\noindent We evaluate our model on four paired datasets. The LOL-v1, LOL-v2-real captured, LOL-v2-synthetic, and SID datasets, which training and testing are split into 485:15, 689:100, 900:100, and 2564:133. We resize the training image to 224x224, and implement our framework with PyTorch on two NVIDIA 4090 GPUs with a batch size of 8. The model is trained using the AdamW optimizer with initial hyperparameters $\beta_1 = 0.9$ and $\beta_2 = 0.999$. Training is conducted for 750 epochs, where the learning rate is initially set to $1 \times 10^{-8}$ and increases to $2 \times 10^{-5}$ over the first 75 epochs during a warmup phase. Subsequently, the learning rate is maintained at $2 \times 10^{-5}$ until the 600th epoch, after which it follows a cosine annealing schedule down to $1 \times 10^{-8}$ towards the end of the training at 750 epochs. To ensure that the enhanced image is perceptually similar to the well-lit ground truth, we employ a perceptual loss \cite{johnson2016perceptual} using the feature maps from a pre-trained VGG-19 network, and we adopt the peak signal-to-noise ratio (PSNR) and structural similarity (SSIM) as the evaluation metrics. 
\begin{equation}
    \mathcal{L}_{vgg} = \sum_{l=1}^{L} \frac{1}{M_l} \|\Phi_l(\hat{I}) - \Phi_l(I_{gt})\|_1
\end{equation}

\begin{table*}[]
\renewcommand{\arraystretch}{1.1}
\centering
\caption{\textbf{Quantitative comparisons} on LOL (v1 \cite{wei2018deep} and v2 \cite{yang2021sparse}) datasets}
\begin{tabular}{l|r|r|c|c|c|c}
\toprule
\textbf{Methods} & \textbf{FLOPs (G)} & \textbf{Params (M)} & \begin{tabular}[c]{@{}c@{}}\textbf{LOL-v1}\\ \textbf{PSNR}\hspace{1em}\textbf{SSIM}\end{tabular} & \begin{tabular}[c]{@{}c@{}}\textbf{LOL-v2-real}\\ \textbf{PSNR}\hspace{1em}\textbf{SSIM}\end{tabular} & \begin{tabular}[c]{@{}c@{}}\textbf{LOL-v2-syn}\\ \textbf{PSNR}\hspace{1em}\textbf{SSIM}\end{tabular} & \begin{tabular}[c]{@{}c@{}}\textbf{SID} \\ \textbf{PSNR} \hspace{1em} \textbf{SSIM}\end{tabular} \\ \midrule
SID \cite{chen2019seeing}        & 13.73                 &7.76  & 14.35  \hspace{1em}      0.436 & 13.24   \hspace{1em}     0.442 & 15.04   \hspace{1em}        0.610 
& 16.97   \hspace{1em}        0.591\\
Zero-DCE \cite{guo2020zero}   & 4.00  &                 0.08 & 14.86   \hspace{1em}     0.667 & 18.06 \hspace{1em}       0.680 & 17.76    \hspace{1em}       0.838 
 & 13.68  \hspace{1em}        0.49\\
RF \cite{kosugi2020unpaired}        & 46.23   &             21.54  & 15.23  \hspace{1em}      0.452 & 14.05    \hspace{1em}    0.458 & 15.97   \hspace{1em}        0.632 
 & 16.44   \hspace{1em}        0.596\\
DeepLPF \cite{moran2020deeplpf}   & 5.86 &                  1.77 & 15.28   \hspace{1em}     0.473 & 14.10    \hspace{1em}    0.480 & 16.02     \hspace{1em}      0.587 
 & 18.07   \hspace{1em}        0.600\\
UFormer \cite{wang2022uformer}   & 12.00   &              5.29  & 16.36  \hspace{1em}      0.771 & 18.82    \hspace{1em}    0.771 & 19.66  \hspace{1em}         0.871 
 & 18.54   \hspace{1em}        0.577\\
RetinexNet \cite{wei2018deep} & 587.47 &              0.84   & 16.77 \hspace{1em}       0.560 & 15.47   \hspace{1em}     0.567 & 17.15    \hspace{1em}       0.798 
 & 16.48   \hspace{1em}        0.578\\
EnGAN \cite{jiang2021enlightengan}     & 61.01  &            114.35   & 17.48  \hspace{1em}      0.650 & 18.23  \hspace{1em}      0.617 & 16.57   \hspace{1em}        0.734 
 & 17.23   \hspace{1em}        0.543\\
RUAS \cite{liu2021retinex}       & 0.83  &               0.003  & 18.23  \hspace{1em}      0.720 & 18.37   \hspace{1em}     0.723 & 16.55     \hspace{1em}      0.652 
 & 18.44   \hspace{1em}        0.581\\
FIDE \cite{xu2020learning}      & 28.51 &                8.62  & 18.27  \hspace{1em}      0.665 & 16.85   \hspace{1em}     0.678 & 15.20      \hspace{1em}     0.612 
 & 18.34   \hspace{1em}        0.578\\
DRBN \cite{yang2021band}      & 48.61 &                5.27  & 20.15    \hspace{1em}    0.830 & 20.29  \hspace{1em}      0.831 & 23.22     \hspace{1em}      0.927 
 & 19.02   \hspace{1em}        0.577\\
KinD \cite{zhang2019kindling}      & 34.99   &              8.02  & 20.86   \hspace{1em}     0.790 & 14.74  \hspace{1em}      0.641 & 13.29   \hspace{1em}        0.578 
 & 18.02   \hspace{1em}        0.583\\
Restormer \cite{zamir2022restormer} & 144.25 &            26.15    & 22.43 \hspace{1em}       0.823 & 19.94  \hspace{1em}      0.827 & 21.41  \hspace{1em}         0.830 
 & 22.27   \hspace{1em}        0.649\\
SNR-Net \cite{xu2022snr}   & 26.35   &              4.01  & \textcolor{blue}{24.61}  \hspace{1em}      0.842 & 21.48   \hspace{1em}     \textcolor{blue}{0.849} & {24.14}   \hspace{1em}        \textcolor{blue}{0.928}
 & \textcolor{blue}{22.87}   \hspace{1em}        0.625\\ 
Retinexformer \cite{cai2023retinexformer} & 15.57 &                1.61 & \textcolor{red}{25.16} \hspace{1em}       \textcolor{blue}{0.845} & \textcolor{blue}{22.80}  \hspace{1em}      0.840 & \textcolor{red}{25.67}   \hspace{1em}        \textcolor{red}{0.930} 
 & \textcolor{red}{24.44}   \hspace{1em}        \textcolor{blue}{0.680}\\ \midrule
\textbf{RSEND (ours)}          & \textbf{17.99}     &            \textbf{0.41} & \textbf{24.18}   \hspace{1em}     \textcolor{red}{\textbf{0.860}} & \textcolor{red}{\textbf{23.92}}  \hspace{1em}      \textcolor{red}{\textbf{0.867}} & \textcolor{blue}{\textbf{24.91}}    \hspace{1em}       \textbf{0.912} 
& \textbf{22.40}   \hspace{1em}        \textcolor{red}{\textbf{0.775}}\\ \bottomrule
\end{tabular}
\label{tab:my_label}
\end{table*}

\subsection{Quantitative Results}
\noindent We compare RSEND with a wide range of state-of-the-art low-light image enhancement networks. Our result significantly outperforms CNN-based SOTA methods on these four datasets, while requiring much less computational and memory cost; the comparison is shown in Table \ref{tab:my_label}.

\noindent Compared with the best CNN-based model DRBN \cite{yang2021band}, our model achieves 4.03, 3.63, 1.69, and 3.38 dB improvements on the LOL-v1, LOL-v2-real, LOL-v2-synthetic, and SID datasets. In addition, our model only consumes 7.8\% (0.41/5.27) parameters and 37 \% (17.99/48.61) FLOPS, which is significantly smaller than many of the other high-performing models, highlighting the efficiency of our architecture. Compared with the SOTA transformer-based model Retinexformer \cite{cai2023retinexformer}, the performance of our model in the LOL-v2-real dataset yields an improvement of 1.12 dB while only consuming 25\% (0.41/1.61) of the parameters. Apart from PSNR, our model achieves SSIM scores of 0.860, 0.867, 0.912, and 0.775 in each dataset, indicating that our model not only accurately restores brightness levels, but also maintains structural integrity and texture details that are crucial for perceptual quality. 

\subsection{Visual and Perceptual Comparisons}
\begin{figure*}[h!]
    \centering
    \begin{subfigure}{.15\textwidth}
        \centering
        \includegraphics[width=\linewidth]{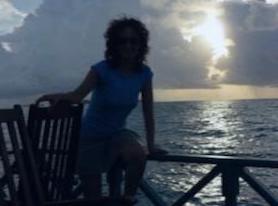}
        \caption{Input}
        \label{fig:input}
    \end{subfigure}%
    \vspace{0.1em}
    \begin{subfigure}{.15\textwidth}
        \centering
        \includegraphics[width=\linewidth]{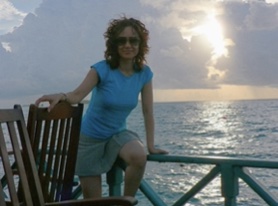}
        \caption{RXformer}
        \label{fig:retinexformer}
    \end{subfigure}
    \begin{subfigure}{.15\textwidth}
        \centering
        \includegraphics[width=\linewidth]{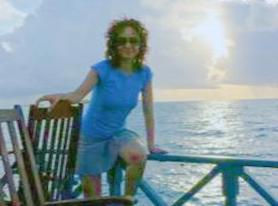}
        \caption{RSEND}
        \label{fig:ELLRetinexformer}
    \end{subfigure}
    \begin{subfigure}{.15\textwidth}
        \centering
        \includegraphics[width=\linewidth]{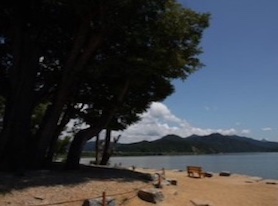}
        \caption{Input}
        \label{fig:input}
    \end{subfigure}%
    \vspace{0.1em}
    \begin{subfigure}{.15\textwidth}
        \centering
        \includegraphics[width=\linewidth]{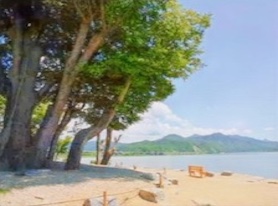}
        \caption{DCC-Net}
        \label{fig:retinexformer}
    \end{subfigure}
    \begin{subfigure}{.15\textwidth}
        \centering
        \includegraphics[width=\linewidth]{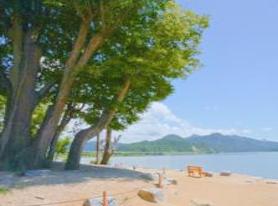}
        \caption{RSEND}
        \label{fig:ELLRetinexformer}
    \end{subfigure}
    \begin{subfigure}{.15\textwidth}
        \centering
        \includegraphics[width=\linewidth]{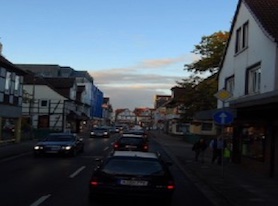}
        \caption{Input}
        \label{fig:input}
    \end{subfigure}%
    \vspace{0.1em}
    \begin{subfigure}{.15\textwidth}
        \centering
        \includegraphics[width=\linewidth]{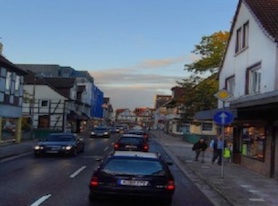}
        \caption{EnGAN}
        \label{fig:retinexformer}
    \end{subfigure}
    \begin{subfigure}{.15\textwidth}
        \centering
        \includegraphics[width=\linewidth]{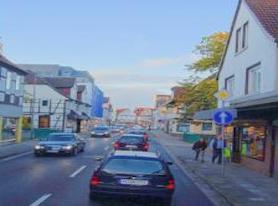}
        \caption{RSEND}
        \label{fig:ELLRetinexformer}
    \end{subfigure}
    \begin{subfigure}{.15\textwidth}
        \centering
        \includegraphics[width=\linewidth]{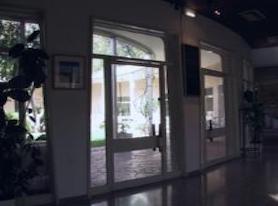}
        \caption{Input}
        \label{fig:input}
    \end{subfigure}%
    \vspace{0.1em}
    \begin{subfigure}{.15\textwidth}
        \centering
        \includegraphics[width=\linewidth]{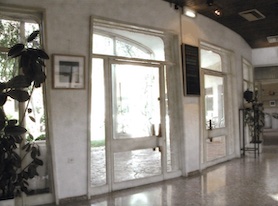}
        \caption{MIRNet}
        \label{fig:retinexformer}
    \end{subfigure}
    \begin{subfigure}{.15\textwidth}
        \centering
        \includegraphics[width=\linewidth]{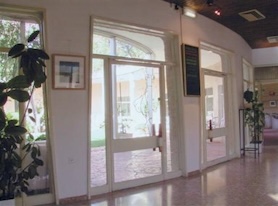}
        \caption{RSEND}
        \label{fig:ELLRetinexformer}
    \end{subfigure}
    \begin{subfigure}{.15\textwidth}
        \centering
        \includegraphics[width=\linewidth]{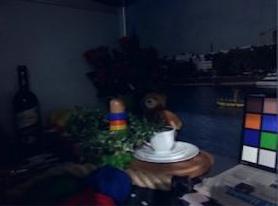}
        \caption{Input}
        \label{fig:input}
    \end{subfigure}%
    \vspace{0.1em}
    \begin{subfigure}{.15\textwidth}
        \centering
        \includegraphics[width=\linewidth]{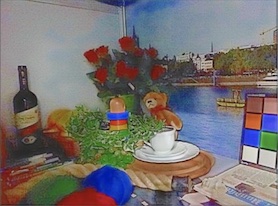}
        \caption{RTXNet}
        \label{fig:retinexformer}
    \end{subfigure}
    \begin{subfigure}{.15\textwidth}
        \centering
        \includegraphics[width=\linewidth]{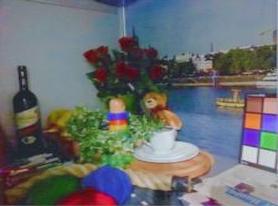}
        \caption{RSEND}
        \label{fig:ELLRetinexformer}
    \end{subfigure}
    \begin{subfigure}{.15\textwidth}
        \centering
        \includegraphics[width=\linewidth]{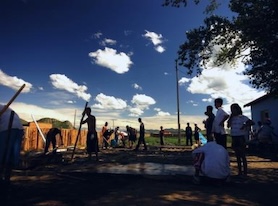}
        \caption{Input}
        \label{fig:input}
    \end{subfigure}%
    \vspace{0.1em}
    \begin{subfigure}{.15\textwidth}
        \centering
        \includegraphics[width=\linewidth]{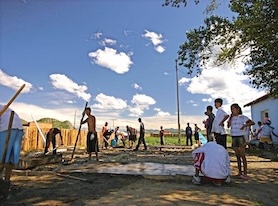}
        \caption{ZeroDCE}
        \label{fig:retinexformer}
    \end{subfigure}
    \begin{subfigure}{.15\textwidth}
        \centering
        \includegraphics[width=\linewidth]{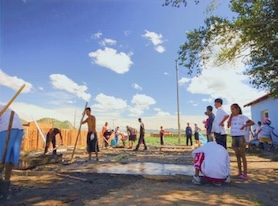}
        \caption{RSEND}
        \label{fig:ELLRetinexformer}
    \end{subfigure}
    \caption{\textbf{Visual comparisons} with Retinexformer \cite{cai2023retinexformer}, DC-Net \cite{zhang2022deep}, EnGAN \cite{jiang2021enlightengan}, MIR-Net \cite{zamir2020learning}, RetinexNet \cite{wei2018deep}, Zero-DCE \cite{guo2020zero} our RSEND performs better.}
    \label{fig:visual-comparison}
\end{figure*}

\noindent Fig. \ref{fig:visual-comparison}. shows the visual comparisons of the low-light image (left), the other model's performance (middle), and our RSEND's performance (right). We can see that our model either makes the image lighter or detects more details, showing the effectiveness of the model in different datasets. 

\subsection{Ablation Study}
\begin{figure*}[h!]
    \centering
    \begin{subfigure}{.18\textwidth}
        \centering
        \includegraphics[width=\linewidth]{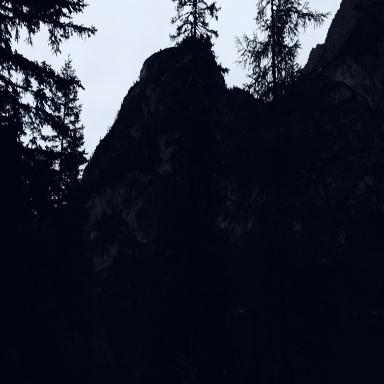}
        \caption{Input}
        \label{fig:input}
    \end{subfigure}%
    \vspace{0.1em}
    \begin{subfigure}{.18\textwidth}
        \centering
        \includegraphics[width=\linewidth]{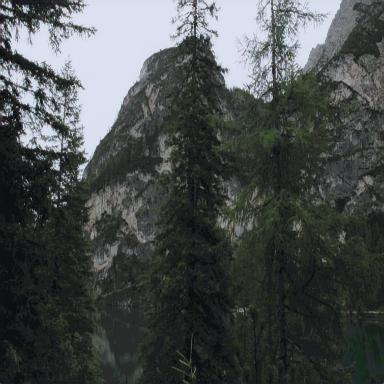}
        \caption{w/o residual}
        \label{fig:residual}
    \end{subfigure}
    \begin{subfigure}{.18\textwidth}
        \centering
        \includegraphics[width=\linewidth]{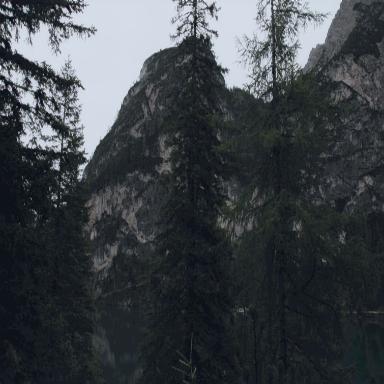}
        \caption{w/o SEB}
        \label{fig:SEB}
    \end{subfigure}
    \begin{subfigure}{.18\textwidth}
        \centering
        \includegraphics[width=\linewidth]{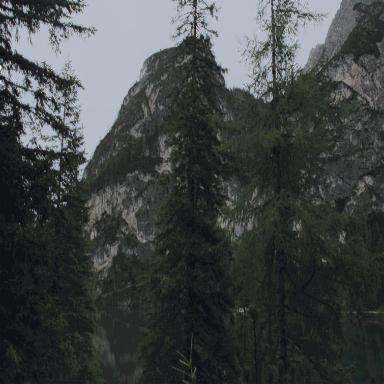}
        \caption{w/o denoise}
        \label{fig:denoise}
    \end{subfigure}
    \begin{subfigure}{.18\textwidth}
        \centering
        \includegraphics[width=\linewidth]{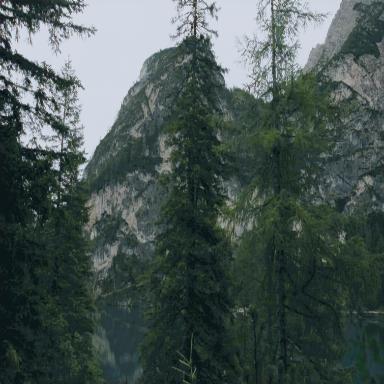}
        \caption{RSEND}
        \label{fig:ELL}
    \end{subfigure}
    \begin{subfigure}{.18\textwidth}
        \centering
        \includegraphics[width=\linewidth]{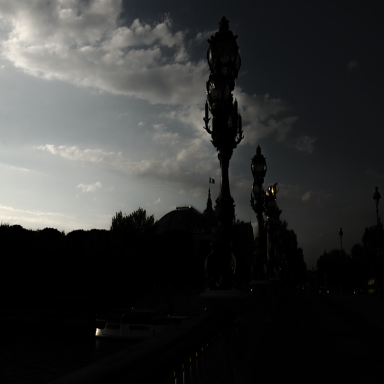}
        \caption{Input}
        \label{fig:input}
    \end{subfigure}%
    \vspace{0.1em}
    \begin{subfigure}{.18\textwidth}
        \centering
        \includegraphics[width=\linewidth]{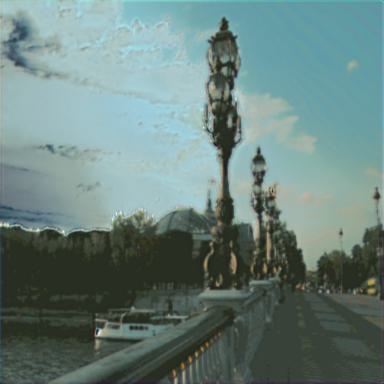}
        \caption{w/o dark}
        \label{fig:nodark}
    \end{subfigure}
    \begin{subfigure}{.18\textwidth}
        \centering
        \includegraphics[width=\linewidth]{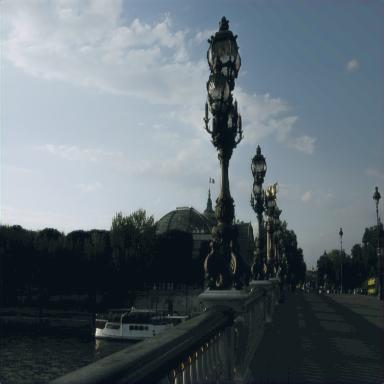}
        \caption{w/o refine}
        \label{fig:norefine}
    \end{subfigure}
    \begin{subfigure}{.18\textwidth}
        \centering
        \includegraphics[width=\linewidth]{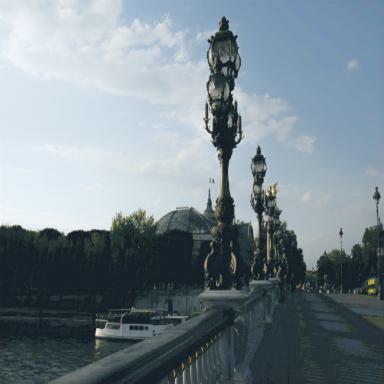}
        \caption{RSEND}
        \label{fig:RSEND2}
    \end{subfigure}
    \caption{Ablation Study of the effect of each model component}
    \label{fig:comparison}
\end{figure*}

\noindent We perform several ablation studies to demonstrate the effectiveness of each part of our network on the LOL-v2-synthetic dataset for its stable convergence. The examples are presented in Fig. \ref{fig:comparison} and Table \ref{tab:ablation}.

\begin{table*}[t]
    \centering
    \caption{Ablation Study of RSEND Components. The table shows the impact of removing each component on PSNR and SSIM performance. \checkmark indicates the presence of the component, while empty denotes its removal.}
    \renewcommand{\arraystretch}{1.2}
    \begin{tabular}{lccccccc}
        \toprule
        \textbf{Method Variation} & \textbf{SEBlock} & \textbf{Dark Region Detection} & \textbf{Residual} & \textbf{Refinement} & \textbf{Denoising} & \textbf{PSNR (dB)} & \textbf{SSIM} \\
        \midrule
        Baseline RSEND             & \checkmark & \checkmark & \checkmark & \checkmark & \checkmark & 24.91 & 0.912 \\
        w/o SEBlock                &  & \checkmark & \checkmark & \checkmark & \checkmark & 20.85 & 0.875 \\
        w/o Dark Region Detection  & \checkmark &  & \checkmark & \checkmark & \checkmark & 21.75 & 0.890 \\
        w/o Residual      & \checkmark & \checkmark &  & \checkmark & \checkmark & 23.06 & 0.902 \\
        w/o Refinement       & \checkmark & \checkmark & \checkmark &  & \checkmark & 22.13 & 0.882 \\
        w/o Denoising              & \checkmark & \checkmark & \checkmark & \checkmark &  & 21.90 & 0.880 \\
        \bottomrule
    \end{tabular}
\label{tab:ablation}
\end{table*}

\begin{figure*}[ht!]
    \centering
    \begin{subfigure}{.18\textwidth}
        \centering
        \includegraphics[width=\linewidth]{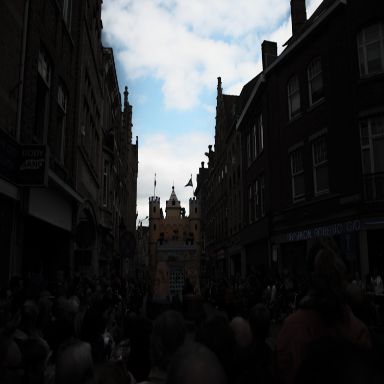}
        \caption{Input}
        \label{fig:input}
    \end{subfigure}%
    \vspace{1em}
    \begin{subfigure}{.18\textwidth}
        \centering
        \includegraphics[width=\linewidth]{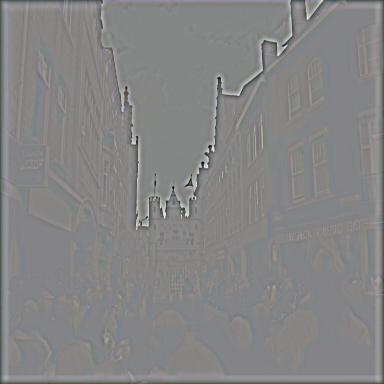}
        \caption{$\mathcal{L}_{col+spa+exp}$}
        \label{fig:combined}
    \end{subfigure}
    \begin{subfigure}{.18\textwidth}
        \centering
        \includegraphics[width=\linewidth]{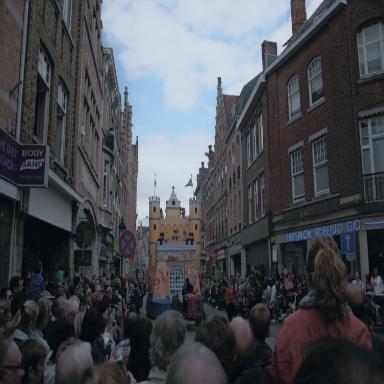}
        \caption{$\mathcal{L}_{Charbonnier}$}
        \label{fig:charon}
    \end{subfigure}
    \begin{subfigure}{.18\textwidth}
        \centering
        \includegraphics[width=\linewidth]{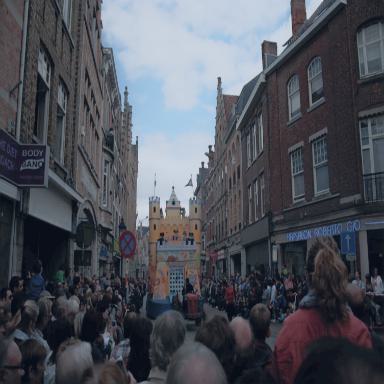}
        \caption{$\mathcal{L}_{comb}$}
        \label{fig:comb_loss}
    \end{subfigure}
    \begin{subfigure}{.18\textwidth}
        \centering
        \includegraphics[width=\linewidth]{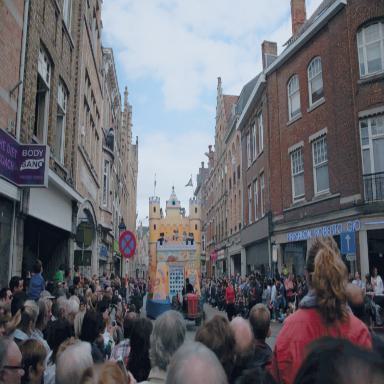}
        \caption{RSEND}
        \label{fig:RSEND_loss}
    \end{subfigure}
    \caption{Ablation Study of the effect of different loss functions}
    \label{fig:comparison_1}
\end{figure*}

\subsubsection{Each Component of the Pipeline}
We conduct ablations to study the effectiveness of each component of the model pipeline. The first is adding the original image at the end. When we just perform element-wise multiplication and denoising, our model yields 23.06 dB in PSNR and 0.902 in SSIM. However, if we add the original image after multiplication, the PSNR and SSIM values are 24.91 dB and 0.912. The difference in Fig. \ref{fig:residual} and Fig. \ref{fig:ELL} shows that with this residual learning-alike algorithm, the model achieves an improvement of 1.85 dB in PSNR and 0.01 in SSIM. The second is the necessity of adding SEBlock. In Fig. \ref{fig:SEB} and Fig. \ref{fig:ELL}, we can clearly see the difference. The output without SEBlock is not well lighted and we can not find visually pleasing details. By adding SEBlock, the PSNR and SSIM values gain an improvement of 4.06 dB and 0.037, suggesting the necessity of using SEBlock. The third is the effect of Denoising after using Retinex theory to get the final image. As we can see in Fig. \ref{fig:denoise} and Fig. \ref{fig:ELL}, even though the output without the denoising phase preserves relatively visually pleasing results, we can still find in some darker areas the details are missing. We get an improvement of 3.15 dB in PSNR and 0.02 in SSIM, which proves the layer's efficacy in mitigating noise and preserving detail. For the rest of the pipelines, as shown in Fig. \ref{fig:nodark}, removing the Dark Region Attention Module results in a significant loss of detail in the darker areas of the image. Even though the image becomes brighter, the color and exposure look unnatural overall, and some parts of the clouds are even black. This demonstrates the module’s effectiveness in enhancing visibility in underexposed regions. As for the image in Fig. \ref{fig:norefine}, without the Refinement Layer, it shows noticeable artifacts and less smooth transitions in lighting, as reflected by a PSNR of 22.13 dB and an SSIM of 0.882, which demonstrates its role in reducing noise and enhancing detail for the enhanced output. Our full RSEND in Fig. \ref{fig:RSEND2} exhibits balanced lighting, enhanced detail, and color accuracy. This result is achieved by integrating all model components, showcasing the synergistic effect that our architectural design aims to accomplish.

\subsubsection{Loss Function Experiment}
Here we supply the results of RSEND trained with various combinations of losses. In Fig. \ref{fig:combined}, it shows the result of the combination of spatial consistency loss, exposure control loss, and color constancy loss, which is inspired by Zero-DCE \cite{guo2020zero}. However, the output is not lighted at all, showing that this combined loss is not suitable for our model. In Fig. \ref{fig:charon}, we use Charbonnier loss\cite{lai2018fast} to train our model and get this output, the loss is a smooth approximation of the L1 loss, to measure the pixel-wise difference between the enhanced image and the ground truth. The image is very close to our RSEND result in Fig. \ref{fig:RSEND_loss}, but it is still not lighted enough with some loss of details in the darker regions. Based on Fig. \ref{fig:RSEND_loss}, we can say that $\mathcal{L}_{vgg}$ plays its role by ensuring that the enhanced image maintains textural and structural similarity to natural images as perceived by the human visual system. Furthermore, we conducted an ablation study to evaluate the performance of all the loss functions combined in \ref{fig:comb_loss}. We set the weights of $\mathcal{L}_{col+spa+exp}$ and $\mathcal{L}_{Charbonnier}$ to be 1 and $\mathcal{L}_{vgg}$ to be 0.5. This decision is based on the need to ensure that the perceptual quality of the enhanced images, as captured by $\mathcal{L}{vgg}$, is sufficiently emphasized without overwhelming the primary objectives of maintaining spatial consistency, proper exposure, and color constancy, as well as minimizing pixel-wise errors. By assigning a lower weight to $\mathcal{L}_{vgg}$, we aim to balance its influence against the other losses. We hoped that the combination would surpass the performance of any single loss function, but it appears that the other losses dragged down the performance of $\mathcal{L}_{vgg}$. This indicates that while $\mathcal{L}_{vgg}$ ensures the textural and structural similarity, the additional losses might have introduced conflicting optimization objectives, thereby diminishing the overall performance.

\section{Conclusion}
\noindent We propose an efficient and accurate CNN-based framework, RSEND, for low-light image enhancement and it can be trained end to end with paired images. RSEND leverages the power of Retinex theory and squeeze and excitation network to significantly enhance the representational power of the network without increasing much computing requirements. 
In RSEND, we make the model understand which parts are darker and require more attention by introducing the dark region detection module. After enhancing, we refine the output, which aims to fine-tune the details. After element-wise multiplication of reflectance and illumination, we add the original low-light image in the end, which serves as residual learning to maintain high similarity. Then we denoise the output image to ensure the final result is not only well-illuminated but also visually pleasing. For all the steps mentioned above, we utilize Squeeze and Excitation Network to better capture the details. The quantitative and qualitative experiments show that our RSEND outperforms all the CNN-based models (by 1.69 dB to 3.63 dB improvements) in PSNR and yields results that are close to or even better than the Transformer-based models (by 1.22 dB to 2.44 dB improvements) while using  3.93–9.78$\times$ fewer parameters compared to the previous state-of-the-art-works.

\bibliographystyle{ieeetr} 
\bibliography{egbib}      
\end{document}